\documentclass[letterpaper, 10 pt, conference]{ieeeconf}
\IEEEoverridecommandlockouts
\usepackage{amsmath}
\usepackage{amssymb}
\usepackage{enumerate}
\usepackage{algorithm}
\usepackage{algpseudocode}
\usepackage{graphicx}
\usepackage{hyperref}
\usepackage{cite}
\usepackage{sidecap}
\usepackage{booktabs}
\usepackage{algorithm}
\usepackage{url}

\usepackage{balance}

\newtheorem{remark}{Remark}

\graphicspath{./images}
\DeclareGraphicsExtensions{.pdf,.eps, .jpg, .png, .PNG}
\title{\LARGE \bf NMPCM: Nonlinear Model Predictive Control on Resource-Constrained Microcontrollers}
\author{Van Chung Nguyen$^1$, Pratik Walunj$^1$, Chuong Le$^1$, An Duy Nguyen$^1$, and Hung Manh La$^1$ 
\thanks{
This work is supported by NASA EPSCoR under grant number 80NSSC24M0141. The views, opinions, findings, and conclusions reflected in this publication are solely those of the authors and do not represent the official policy or position of NASA.}
\thanks{$^1$are with the Advanced Robotics and Automation (ARA) Lab, Department of Computer Science and Engineering, University of Nevada, Reno, NV 89557, USA.}
\thanks{Available source code: \url{https://github.com/aralab-unr/NMPCM}}
\thanks{Corresponding author: Hung La, email: {\tt\small hla@unr.edu}}%
}
\begin{document}
\maketitle
\setlength{\textfloatsep}{0pt}
\begin{abstract} 
Nonlinear Model Predictive Control (NMPC) is a powerful approach for controlling highly dynamic robotic systems, as it accounts for system dynamics and optimizes control inputs at each step. However, its high computational complexity makes implementation on resource-constrained microcontrollers impractical. While recent studies have demonstrated the feasibility of Model Predictive Control (MPC) with linearized dynamics on microcontrollers, applying full NMPC remains a significant challenge. This work presents an efficient solution for generating and deploying NMPC on microcontrollers (NMPCM) to control quadrotor UAVs. The proposed method optimizes computational efficiency while maintaining high control accuracy. Simulations in Gazebo/ROS and real-world experiments validate the effectiveness of the approach, demonstrating its capability to achieve high-frequency NMPC execution in real-time systems. The code is available at: \url{https://github.com/aralab-unr/NMPCM}.
\end{abstract}
\section{Introduction}
Over the past decade, Model Predictive Control (MPC) has gained significant interest in both industrial and academic domains due to its ability to solve online optimization problems at each time step, ensuring optimal performance \cite{nguyen2024tinympc, manchester2019contact,vukov2012experimental,wensing2023optimization}. On the one hand, with the increasing speed of convex quadratic programming solvers, Linear MPC (LMPC) has been successfully applied to a wide range of systems \cite{nguyen2024tinympc, pawlowski2022linear, liao2021analysis}. One effective approach for solving convex quadratic programming problems is the Alternating Direction Method of Multipliers (ADMM). This method enables the direct computation of solutions for LMPC, facilitating high-frequency implementations. The effectiveness of this approach has been demonstrated in TinyMPC \cite{nguyen2024tinympc}, which achieves real-time performance even on resource-constrained microcontrollers. On the other hand, Nonlinear Model Predictive Control (NMPC) has also attracted significant research attention \cite{romero2022model, rathai2019gpu, meduri2023biconmp, santos2022nonlinear}. However, due to the inherent complexity of NMPC, most recent implementations have been limited to systems with slower dynamics. Achieving high-frequency, real-time performance often requires the use of off-board computers or powerful on-board computing resources.

\begin{figure*}
\centering
\centerline{\includegraphics[width=0.95\linewidth]{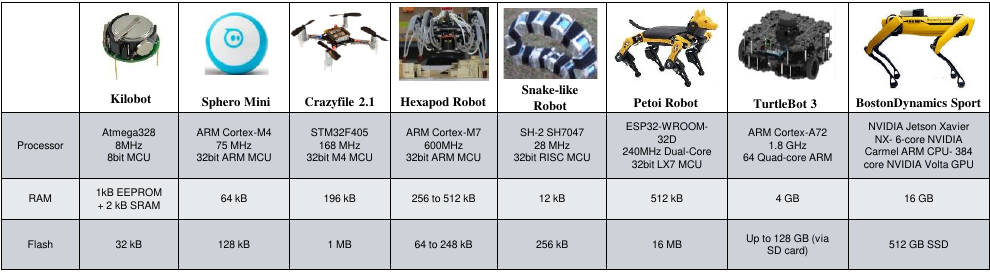}}
\caption{A detailed hardware comparison of various robot platforms, highlighting their processor, RAM, and flash memory. At the smallest scale, the Kilobot \cite{rubenstein2012kilobot} exemplifies a microrobot that utilizes constrained 8-bit microcontrollers to execute pre-programmed motions. At slightly larger scales, platforms such as the Sphero Mini \cite{SpheroMini}, Crazyflie 2.1 \cite{crazyflie}, Hexapod robot \cite{bachega2023flexibility}, Snake-like robot \cite{ohashi2010loop}, and Petoi Robot \cite{Petoi} employ more advanced 32-bit microcontrollers for motion planning and control. At the full scale, platforms like the TurtleBot 3 \cite{amsters2020turtlebot} and Boston Dynamics' Spot \cite{bostondynamic} leverage onboard computers such as the Raspberry Pi 4 and NVIDIA Jetson Xavier, enabling high-performance computing and the execution of complex tasks. In this research, we focus on tiny-scale platforms that utilize resource-constrained microcontrollers to achieve full task control and solve optimization problems. While these microcontrollers are capable of limited onboard computation, they exhibit significantly lower processor speeds, RAM, and flash memory compared to full-scale robotic systems.}
\label{robotcontroller}
\vspace{-10pt}
\end{figure*}

In parallel, there has been increasing interest in the development of small, cost-effective robots capable of navigating confined spaces, rendering them highly suitable for a wide range of applications. These include emergency search and rescue missions \cite{mcguire2019minimal} as well as routine inspection and maintenance of infrastructure and equipment \cite{duisterhof2021sniffy}. Such robots typically rely on low-power, resource-constrained microcontrollers (MCUs) to manage their computational tasks \cite{nguyen2024tinympc, giernacki2017crazyflie}. However, due to these resource limitations, most miniature robots depend on off-board computers to execute complex tasks and receive commands via receivers \cite{adajania2023amswarm, varshney2019deepcontrol, lambert2019low, luis2020online, xi2021gto}. As illustrated in Figures~\ref{robotcontroller}, the constraints imposed by limited RAM, flash memory, and processor speed on MCUs make the implementation of complex tasks impractical compared to off-board computers or more powerful on-board computing systems, such as the Jetson Nano or LattePanda.

In addition to the computational challenges of NMPC, this limitation poses a significant barrier to deploying NMPC on resource-constrained microcontrollers, which are increasingly being utilized in embedded systems, robotics, and other applications. To achieve highly efficient implementations of algorithms for solving NMPC problems online within these constraints, code generation has emerged as a promising solution. For instance, tools like AutoGenU \cite{houska2011acado} generate source code implementing the continuation/GMRES method, while IPOPT, used in CasADi \cite{Andersson2019}, and the OSQP solver for convex quadratic programming \cite{osqp} offer additional capabilities. Similarly, the software package CVXGEN \cite{mattingley2012cvxgen} enables users to generate customized interior-point solvers for small-scale linear and quadratic programming problems, and the qpOASES solver \cite{Ferreau2014} employs active-set algorithms, which form a second class of suitable QP solvers commonly encountered in MPC. Inspired by the ACADO Code Generation tool, which successfully implements nonlinear model predictive control on resource-limited platforms such as the Raspberry Pi \cite{adhau2019implementation}, this paper presents a method for implementing NMPC on resource-constrained microcontrollers. Furthermore, the initial solution for tracking the reference trajectory of quadrotor UAVs is handled via a Cascaded PID controller, which provides an initial guess to enhance the performance of the NMPCM. We also provide simulations in Gazebo/ROS based on code generated using ACADO, demonstrating the feasibility of high-frequency NMPCM at rates of up to 1 kHz.
In comparison with related works, the main contributions of this paper can be outlined as follows:
\begin{itemize}
    \item A novel framework for implementing generated source code of NMPC on resource-constrained microcontrollers (NMPCM), incorporating an initial solution guess to enhance the performance and convergence of the NMPC solver.
    \item Open-source C++ code for NMPC using ACADO-generated code and the qpOASES solver, capable of running in real-time on microcontrollers.
    \item Open-source simulation code for high-frequency NMPCM, achieving execution rates of up to 1 kHz in Gazebo/ROS 2. Experimental results demonstrate the feasibility of NMPCM and other complex tasks on a single resource-constrained microcontroller.
\end{itemize}
The paper is organized as follows: Section II introduces the kinematic model of quadrotor UAVs. Section III derives the NMPC formulation with a virtual reference of the Cascaded PID controller and details the code generation process using ACADO and the qpOASES solver for implementation on microcontrollers (MCUs). Section IV presents the Gazebo/ROS 2 simulation results, benchmarking studies, and hardware experiments conducted on a single MCU for quadrotor UAVs. Finally, Section V concludes the manuscript with a discussion and future works.
\section{Modelling}

In this research, we consider the full kinematic model of the UAV as follows:
\begin{align} 
& m\ddot{p}  = (s_{\psi} s_{\phi}+c_{\psi} s_{\theta} c_{\phi}) U_1  \label{kine1} \\ 
& m\ddot{q}  =     (s_{\psi} s_{\theta} c_{\phi}-c_{\psi} s_{\phi}) U_1   \\
& m\ddot{r}  =  c_{\theta} c_{\phi} U_1 -mg \\
& I_{xx} \ddot{\phi}  = \dot{\theta}\dot{\psi} (I_{yy}-I_{zz}) + U_2 \\
& I_{yy} \ddot{\theta}  =\dot{\psi}\dot{\phi}(I_{zz}-I_{xx}) + U_3 \\
& I_{zz} \ddot{\psi}  = \dot{\phi}\dot{\theta}(I_{xx}-I_{yy}) + U_4 \label{fullmodeluavend}
\end{align}
where $p,q,r$ denote the position of the UAV; $\phi,\theta,\psi$ denote the rotation of the UAV, $I_{xx}, I_{yy}, I_{zz}$ are the rotational inertias; and $c, s$ are the shorthand form of cosine and sine, respectively. The control inputs are $U_1, U_2, U_3, U_4$. 
For the compact form, the kinematic model of the Quadrotor UAV is expressed as follows:
\begin{align} \label{nonlinearmodel}
    \dot{\mathbf{x}}(t) = \mathbf{f}(\mathbf{x}(t), \mathbf{u}(t))
\end{align}
where $\mathbf{x}(t) = [p \ q \ r \ \phi \ \theta \ \psi \ \dot{p} \ \dot{q} \ \dot{r} \  \dot{\phi} \ \dot{\theta} \ \dot{\psi}]^T$, and the control law is $\mathbf{u}(t) = [U_1 \ U_2 \ U_3 \ U_4]^T$.
\section{Methodology}
\subsection{NMPC formulation}
In this paper, we consider the nonlinear MPC in the following form:
\begin{align}
\min_{\mathbf{x}(\cdot), \mathbf{u}(\cdot)} & J (\mathbf{x}(\cdot),\mathbf{u}(\cdot) ) = \|\mathbf{x}(t+T) - \mathbf{x}^{\mathrm{r}}(t+T)\|_R^2 \label{nmpcformu} \\
&  + \int_t^{t+T} (\|\mathbf{x}(\tau) - \mathbf{x}^{\mathrm{r}}(\tau)\|_P^2 + \|\mathbf{u}(\tau) - \mathbf{u}^{\mathrm{r}}(\tau)\|_Q^2)\,d\tau  \notag  \\
\text{s.t.} \quad &  \notag \\
& \mathbf{x}(t) = \mathbf{x}_t \label{nmpcinitial} \\
&\dot{\mathbf{x}}(\tau) = \mathbf{f}(\mathbf{x}(\tau), \mathbf{u}(\tau)) \quad \text{for all } \tau \in [t, t+T] \\
&\underline{\mathbf{u}} \leq \mathbf{u}(\tau) \leq \overline{\mathbf{u}} \quad \text{for all } \tau \in [t, t+T] \\
&\underline{\mathbf{x}} \leq \mathbf{x}(\tau) \leq \overline{\mathbf{x}} \quad \text{for all } \tau \in [t, t+T]
\end{align}
Here, \( \mathbf{x} \) and \( \mathbf{u} \) represent the differential state and control input, respectively; \( t \) denotes the current running time of the NMPC. \( \mathbf{x}_t \) represents the initial state, which is the current state of the robot, and \( T \) denotes the length of the prediction horizon. The upper and lower constraints for the control input and differential states are \( \underline{\mathbf{u}} \), \( \overline{\mathbf{u}} \), \( \underline{\mathbf{x}} \), and \( \overline{\mathbf{x}} \), respectively. The cost function objective is weighted with positive-definite matrices \( P \), \( Q \), and \( R \). Additionally, the variables \( \mathbf{x}^{\mathrm{r}} \) and \( \mathbf{u}^{\mathrm{r}} \) denote the reference for the system, and the term \( \|\mathbf{x}(t+T) - \mathbf{x}^{\mathrm{r}}(t+T)\| \) represents the terminal cost for the NMPC. Note that this problem can be identified as a parameterized optimal control problem (OCP), where the input is the initial condition, which is the current state of the robot \( x_t \). The NMPC solves the OCP at each time step based on the initial condition and provides the control input for the system. More details and setups about the NMPC controller can be found in \cite{vukov2012experimental}. Solving this problem yields the optimal solution for the controller. However, under the influence of gravity, the control reference for quadrotor UAVs at the equilibrium point is non-zero, as the thrust generated by the motors must continuously counteract gravity. Providing an appropriate reference input for the NMPC will improve the robot's performance and also serve as an initial guess for the solvers. For the warm start to generate the control reference in \eqref{nmpcformu}, we propose using a Cascaded PID controller. The NMPC will then utilize the control input from the Cascaded PID controller to optimize the robot's performance while adhering to the constraints at each time step.

\begin{figure} 
\centering
{\resizebox*{6.5cm}{!}{\includegraphics{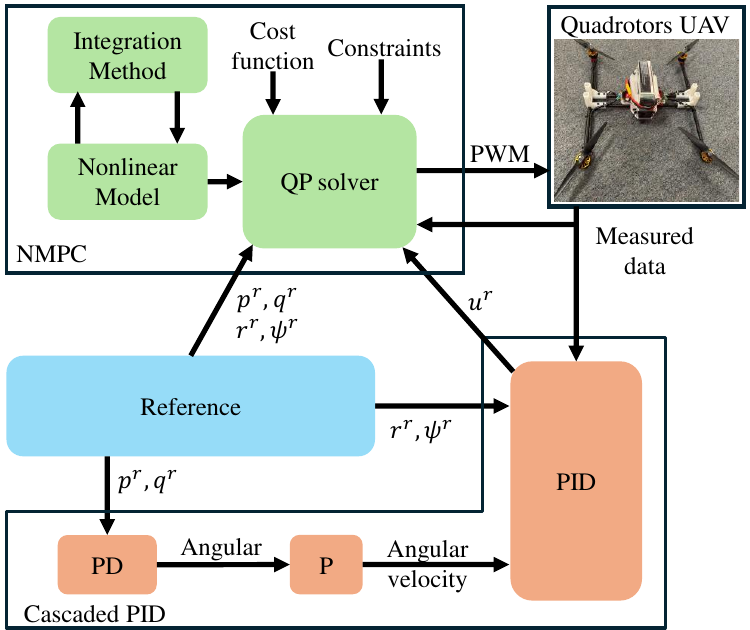}}}\hspace{5pt}
\caption{The control diagram of the NMPC-PID.} 
\label{schemecontrolnmpc}
\end{figure}

\subsection{ACADO code generation and online QP solver qpOASES}
In this section, we provide a brief overview of C++ ACADO code generation and the qpOASES solver. An example of the generated code, along with instructions on how to generate the code and include qpOASES, can be found in \cite{houska2011auto}.

ACADO code generation is designed to address nonlinear optimal control problems, which are typically formulated as general quadratic programs (QPs). To solve these nonlinear optimal control problems, ACADO discretizes the nonlinear dynamics using the multiple shooting method \cite{bock1984multiple}. This approach transforms the problem into a nonlinear programming problem (NLP), which can then be solved using the sequential quadratic programming (SQP) method. Furthermore, within the tracking term of the cost function in \eqref{nmpcformu}, ACADO employs the Gauss-Newton method to approximate the Hessian matrices. This enables the generation of real-time iterations, where only one SQP step is required per time step. To solve the SQP problem, ACADO incorporates two tailored algorithms: CVXGEN \cite{mattingley2012cvxgen} and qpOASES \cite{Ferreau2014}. CVXGEN utilizes a primal-dual interior-point solver, and the exported code is implemented in plain C, using only static memory. This results in relatively constant computation times for each time step. On the other hand, qpOASES employs active-set algorithms, which represent a second class of QP solvers. For ACADO code generation, modifications such as hard-coded dimensions and static memory allocation are utilized. While the computation time for active-set methods is harder to predict due to its dependence on the number of active-set changes, it benefits from dedicated hot-starting capabilities and is generally faster than interior-point iterations.

Given these advantages, we employ the qpOASES solver in this paper. The implementation of NMPC on resource-constrained microcontrollers requires minimal static memory consumption and a fast algorithm to ensure real-time performance. 

\begin{remark}
    Considering the NMPC problem in \eqref{nmpcformu} with a differential solver (such as the fourth-order Runge–Kutta method or the ODE45 numerical integrator) for the robot dynamics, the nonlinear optimal problem becomes a QP problem that can be solved using the active-set method of the qpOASES solver to achieve the solution for the system.
\end{remark}

\begin{figure} 
\centering
{\resizebox*{8.cm}{!}{\includegraphics{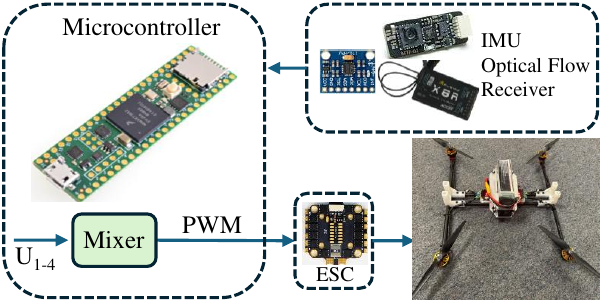}}}\hspace{5pt}
\caption{Scheme of the NMPC on MCUs (Teensy 4.1).} 
\label{schemenmpc}
\end{figure}
\section{Results}
To demonstrate the efficiency of the proposed method, we test it through both simulations in Gazebo/ROS 2 and experiments on a resource-constrained microcontroller. Moreover, we benchmark our embedded controller against a controller that uses state-of-the-art Model predictive contouring control \cite{romero2022model} (MPCC) and NMPC solved by CasADi \cite{Andersson2019} to show the efficacy of our embedded solver. The results show that our controller not only achieves a high frequency for the NMPC but also tracks the trajectories while satisfying the control input constraints.
\subsection{Simulation Results}
\begin{figure} [H]
\centering
{\resizebox*{8.0 cm}{!}{\includegraphics{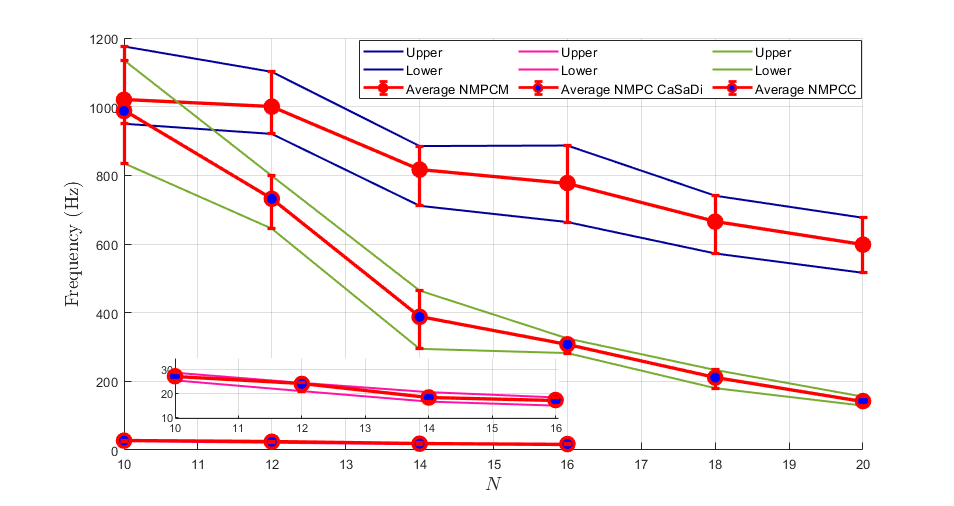}}}
\vspace{-10pt}
\caption{Solving time of the proposed method, MPCC using ACADO code generation, and NMPC using CaSaDi (all employing RK4 for integration). The prediction horizons range from 10 to 20, with the integral step set to 5. While CaSaDi successfully solves the problem, the quadrotor fails to track the reference trajectory when the prediction horizon exceeds 16. The proposed method outperforms others in solving time while ensuring the tracking performance of the robot and satisfying the constraints of the NMPC problem.} 
\label{solvetime}
\end{figure}

\begin{figure} [H]
\vspace{-20pt}
\centering
{\resizebox*{4.25 cm}{!}{\includegraphics{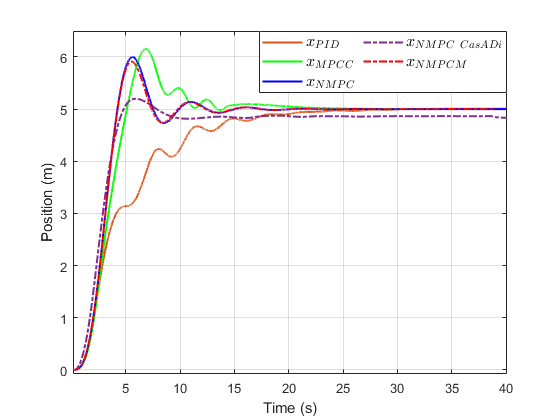}}}
{\resizebox*{4.25 cm}{!}{\includegraphics{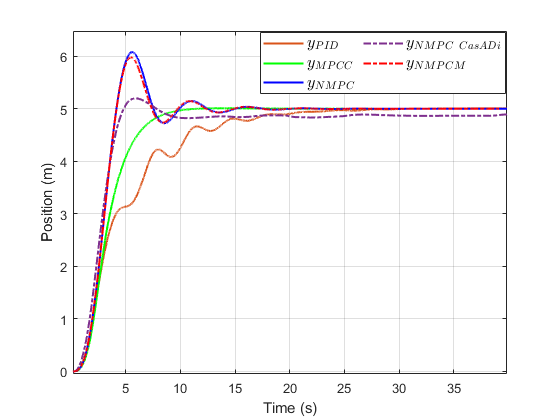}}}
\vspace{-20pt}
\caption{Tracking performance along the x and y-axis.} 
\label{xysimu}
\end{figure}

\begin{figure} [H]
\centering
{\resizebox*{4.25 cm}{!}{\includegraphics{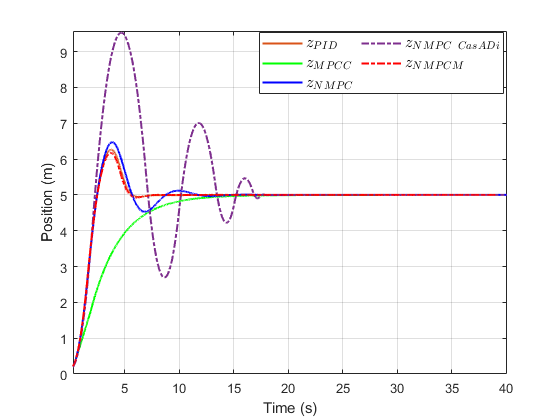}}}
{\resizebox*{4.25 cm}{!}{\includegraphics{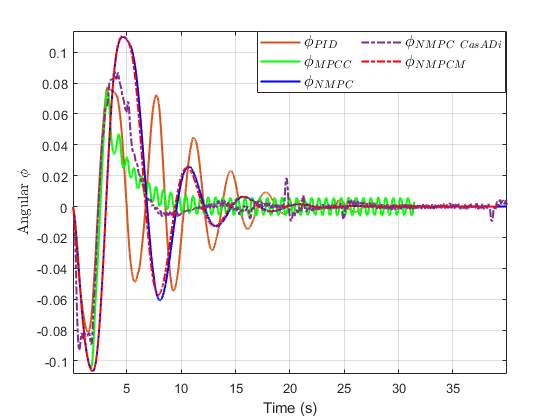}}}
\vspace{-20pt}
\caption{Tracking performance along the z-axis and angular $\phi$} 
\label{zphisimu}
\end{figure}

\begin{figure} [H]
\vspace{-20pt}
\centering
{\resizebox*{4.25 cm}{!}{\includegraphics{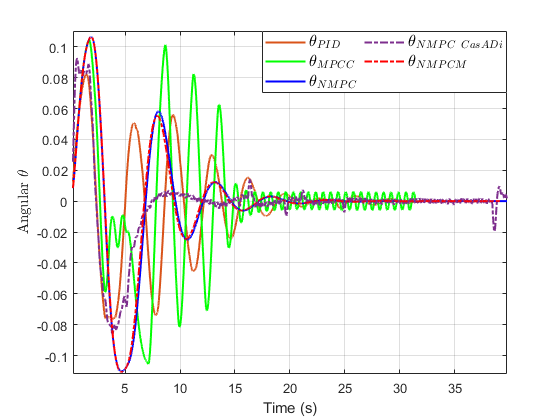}}}
{\resizebox*{4.25 cm}{!}{\includegraphics{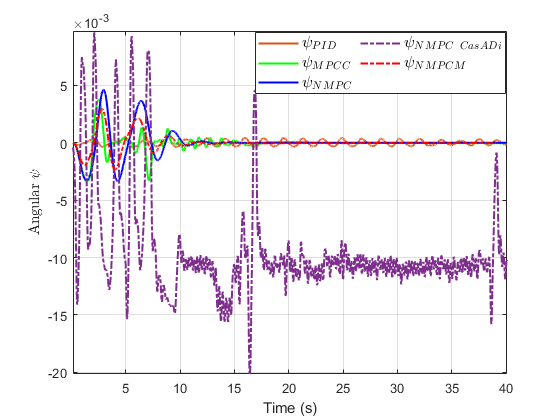}}}
\vspace{-20pt}
\caption{The angular $\theta$, and $\psi$.} 
\label{thetapsisimu}
\end{figure}

\begin{figure} [H]
\vspace{-20pt}
\centering
{\resizebox*{4.25 cm}{!}{\includegraphics{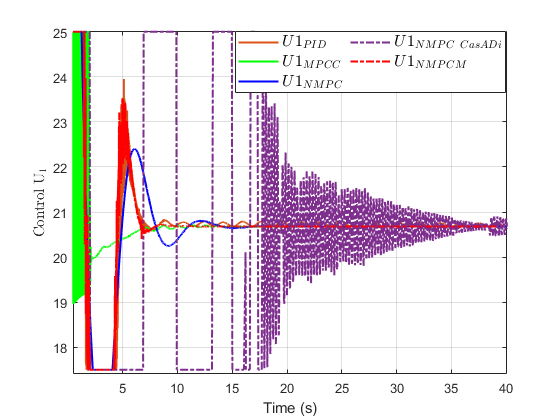}}}
{\resizebox*{4.25 cm}{!}{\includegraphics{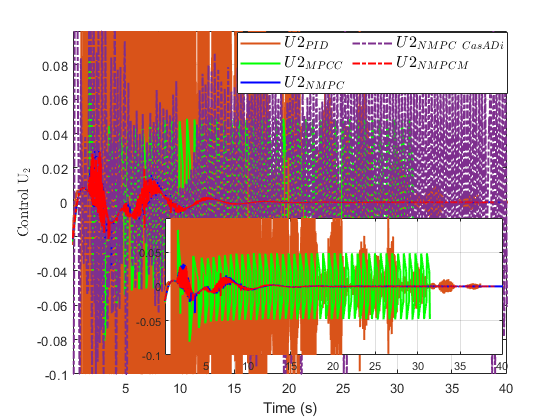}}}
\vspace{-20pt}
\caption{Control law $U_1$ and $U_2$.} 
\label{u1u2simu}
\end{figure}

\begin{figure} [H]
\vspace{-20pt}
\centering
{\resizebox*{4.25 cm}{!}{\includegraphics{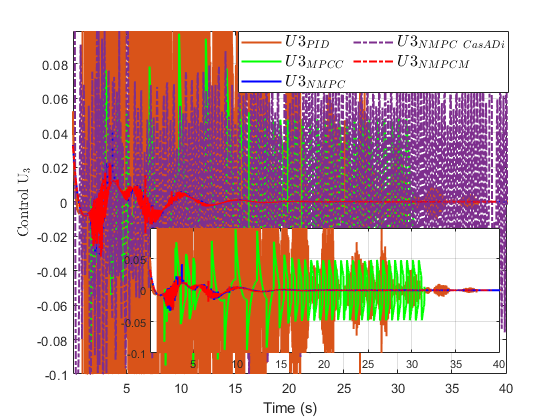}}}
{\resizebox*{4.25 cm}{!}{\includegraphics{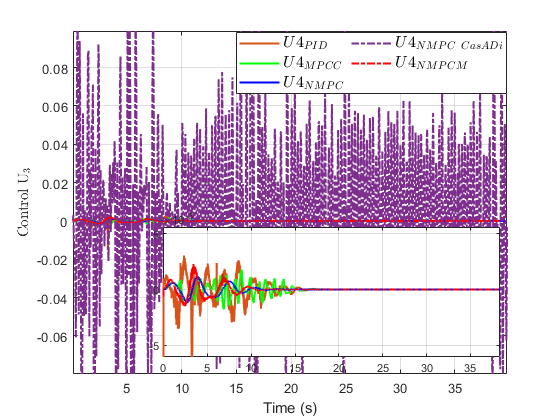}}}
\vspace{-20pt}
\caption{Control law $U_3$ and $U_4$.} 
\label{u3u4simu}
\end{figure}

The quadrotor models are designed in SolidWorks, and URDF files are exported from these models to determine the robot's parameters. The robot and control parameters are configured as shown in Table~\ref{robotpara}. The initial position of the flying robot is set to $[x_0, y_0, z_0] = [0, 0, 0.15]$ meters, and the desired position is set to $[5, 5, 5]$ meters. The simulation results are presented in Figures~\ref{solvetime} to \ref{u3u4simu}. The simulations are performed on a computer equipped with an Intel i7-9700F CPU, 16GB RAM, and an NVIDIA GeForce RTX 2060 GPU. The solving time of the NMPCM and MPCC using embedded ACADO and CaSaDi is illustrated in Figure~\ref{solvetime}. The proposed method achieves a solving frequency of up to 1 kHz, while the maximum frequency of CaSaDi is limited to 30 Hz. For tracking performance, the position and angular states of the robot are shown in Figures~\ref{xysimu}--\ref{thetapsisimu}. Although the standard NMPC (ACADO) and the NMPCM exhibit larger overshoots than NMPC (CaSaDi) along the x- and y-axes, they do not experience steady-state errors. In contrast, NMPC (CaSaDi) shows a significant overshoot in the drone's height. Additionally, the Cascaded PID controller can stabilize the drone at high frequencies. However, the control input may exhibit fluctuations due to its high-frequency operation. The tracking performance of the MPCC shows no overshoot in the y and z-axes since the reference is solved at each time step alongside the NMPC. However, when the robot reaches the desired point, the MPCC exhibits fluctuations at the equilibrium points. The control laws are illustrated in Figures~\ref{u1u2simu}--\ref{u3u4simu}.

Finally, we benchmark the performance of the controllers in Table~\ref{benchmarksimu}. The benchmark terms include settling time, overshoot, maximum control signal, Integral Square Error (ISE), Integral of Time Multiplied Square Error (ITSE), Integral Absolute Error (IAE), and Integral of Time Multiplied Absolute Error (ITAE).
\begin{table}[H] 
    \centering
    \caption{The system parameter and controller constraints}
    \vspace{-5pt}
    \begin{tabular}{|c|c|}
    \hline
     \label{robotpara}
    \textbf{Property} & \textbf{Customize  Quad} \\ \hline
    $m$ [kg] & 2.11 \\ \hline
    $l$ [m] & 0.159 \\ \hline
    $\text{diag}(J)$ [gm$^2$] & [0.0785, 0.0785, 0.105] \\ \hline
    $[U_{1\min}, U_{1\max}]$ [N] & [17.5, 25.0] \\ \hline
    $[U_{2\min}, U_{2\max}]$ [Nm] & [-0.1, 0.1] \\ \hline
    $[U_{3\min}, U_{3\max}]$ [Nm]  & [-0.1, 0.1] \\ \hline
    $[U_{4\min}, U_{4\max}]$ [Nm] & [-0.1, 0.1] \\ \hline
    \end{tabular}
     \vspace{-15pt}
\end{table}
\begin{table*} [h]
\centering
\caption{The control performance benchmark}
{\begin{tabular}
{|c|c|c|c|c|c|c|c|} \hline
\label{benchmarksimu}
\bf{Methods}&{Setting Time 5\% (s)}& Overshoot (\%)& $U_{\text{Max}}$ & ITAE & IAE & ISE & ITSE   \\ \hline
\text{NMPC (ACADO)}    & 8.799 & 29.45 & [25, 0.0331, 0.0415, 0.0011] & 27.5078 & 11.4523 & 31.6083 & 35.1319  \\ 
\text{NMPC (CasADi)}     & 16.607 & 90.58 & [25, 0.1, 0.1, 0.1] & 188.604 & 32.4622 & 98.0721 & 436.9455 \\ 
\text{Cascaded PID}  & 14.099  & 25.23 & [25, 0.1, 0.1, 0.0030] & 121.7324 & 26.1225 & 75.3648 & 170.3765  \\  
\text{MPCC \cite{romero2022model}}  & 10.627  & 24.21 & [25, 0.0955, 0.0818, 0.0034] & 49.2465 & 16.8834 & 47.6453 & 75.0222  \\  
\text{NMPCM (our)}  & 8.612 & 23.46 & [25, 0.0295, 0.0359, 0.0023] & 14.6793 & 9.4495 & 29.6286 & 26.3615 \\  \hline
\end{tabular}}
\label{parameter}
\end{table*}

\subsection{Experiment Results}
\begin{figure} [H]
\centering
{\resizebox*{7.5 cm}{!}{\includegraphics{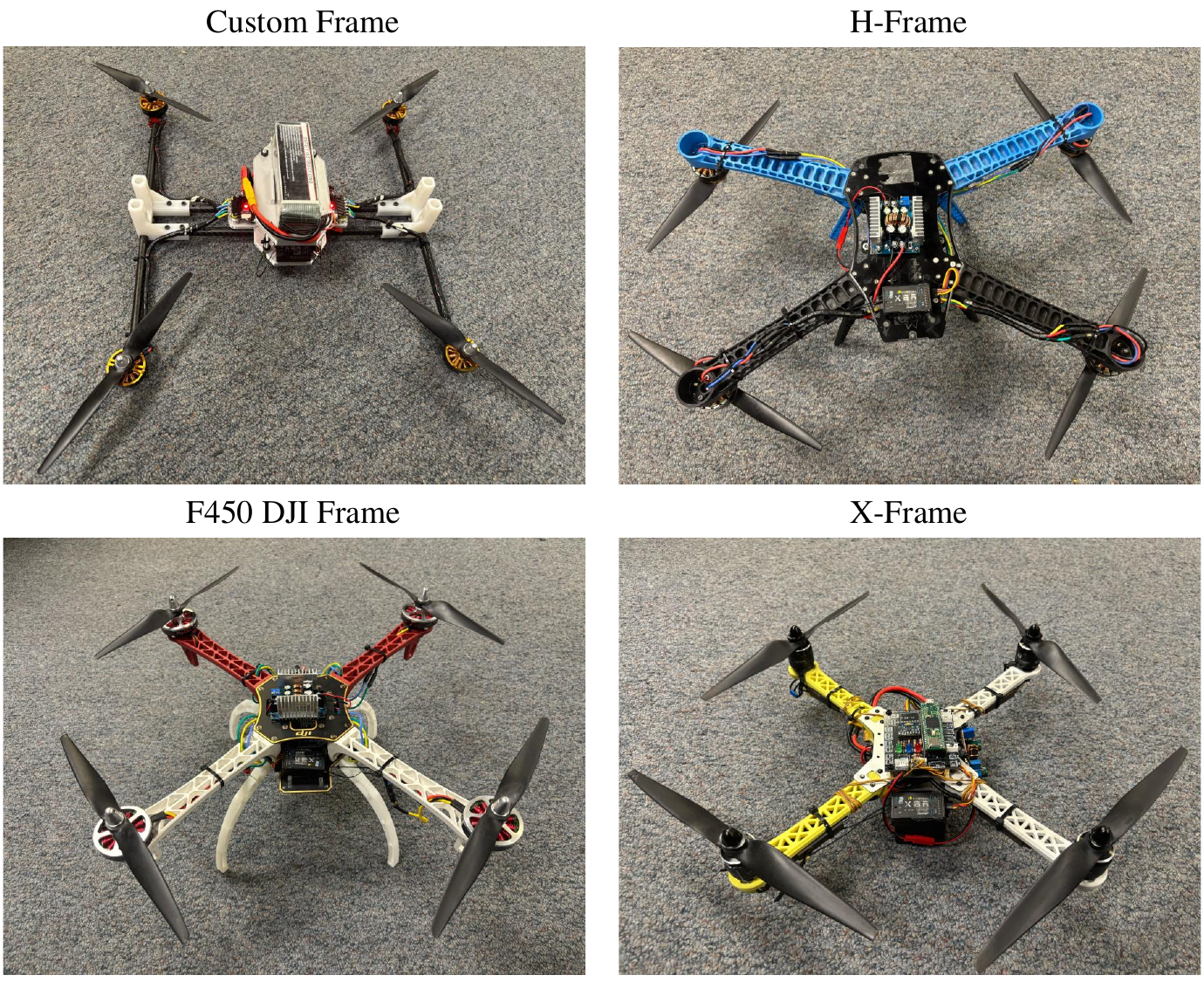}}}
\caption{Different types of quadrotor UAV frames.}
\vspace{-10pt}
\label{droneframe}
\end{figure}

We verified the efficiency of our control method for real-time execution on the resource-constrained Teensy 4.1 microcontroller and implemented our flight controller across various quadrotor platforms. We tested our flight controller on four types of platforms, as shown in Figure~\ref{droneframe}. Additionally, we evaluated the NMPCM solving time on the Teensy 4.1, demonstrating that our approach can achieve high-frequency execution on resource-constrained hardware, despite the computational demands of NMPC. The implementation of our controller is based on Algorithm~\ref{alg:nmpcpid}.
\begin{figure*} 
\centering
{\text{a.}\resizebox*{8.25 cm}{!}{\includegraphics{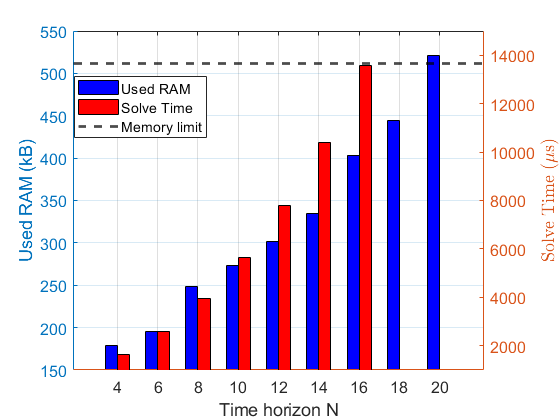}} }
{\text{b.}\resizebox*{8.25 cm}{!}{\includegraphics{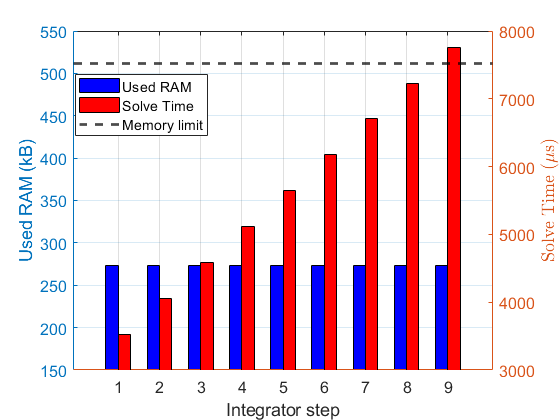}}}
\caption{A benchmark of memory usage and solution time for each interaction of embedded NMPC on the Teensy 4.1 development board (ARM Cortex-M7 running at 600 MHz with 32-bit floating-point support, 7.75 MB of flash, 512 kB of tightly coupled RAM1, and 512 kB of RAM2). In this research, we limited the memory usage to 512 kB for all variables and programmed the system to use only RAM1. At a, with the integrator step set to 5, the time horizon limit is 18, beyond which the microcontroller fails to solve the optimization problem. At b, with the time horizon set to 10, the memory consumption remains consistent regardless of the integrator step. However, as the integrator step increases, the solution time also increases. This represents a trade-off between the accuracy and the solution time of the NMPCM.} 
\label{benchmark}
\end{figure*}
\begin{algorithm} [H]
\caption{NMPCM}\label{alg:nmpcpid}
    \begin{algorithmic}
    \Function{NMPCM\underline{{ }{ }}solver}{}
        \While {\text{(Threshold)}}
        \State \texttt{//Received \& processed sensor data}
        \State $ x_s \gets \text{IMU \& Optical flow data}$
        \State $ x_t \gets \text{Kalman Filter} (x_s)$
        
        \State \texttt{//Calculated the reference}
        \State $x_\text{r} \gets \text{Get radio receiver data}$
        \State $u_\text{r} \gets \text{Cascaded PID}(x_t)$
        
        \State \texttt{//NMPCM solver}
        \State $\text{Update the reference $u_{\text{r}}$ on \eqref{nmpcformu}}$
        \State $\text{Update the initial states $x_t$ on  \eqref{nmpcinitial}}$
        \State $u_{1-4} \gets \text{First step on NMPC solution}$
        
        \State \texttt{//Actuator}
        \State $\text{Convert $u_{1-4}$ into the PWM range and mix}$
        \State $\text{Send the PWM to the ESC}$
        \EndWhile
    \EndFunction
    \end{algorithmic}
\end{algorithm}
\subsubsection{Hardware setup}
The flight controller is written in Arduino IDE, and the source code is provided at \url{https://github.com/aralab-unr/NMPCM}. The custom quadrotor runs on the Teensy 4.1 microcontroller, with ACADO code generation in C++ using the qpOASES solver. The generated code is then embedded in the microcontroller. The drone's angular and position data are obtained from the MPU 6050 and the Micoair MTF-02 optical flow and LiDAR sensor. We use the FrSky X8R radio receiver to send the desired angular commands to the quadrotor UAV.
\subsubsection{Microcontroller solving time benchmark}
The solving time and memory usage when executing the NMPC on the Teensy 4.1 are shown in Figure~\ref{benchmark}. The Teensy 4.1 features an ARM Cortex-M7 microcontroller operating at 600 MHz, with 7.75 MB of flash memory and 512 kB of RAM. Therefore, the memory usage must not exceed this RAM limit. Moreover, the solving time must be fast enough for real-time implementation. The parameters of the embedded NMPCM are set the same as in the simulation.
\begin{figure} [H]
\centering
{\resizebox*{4.25 cm}{!}{\includegraphics{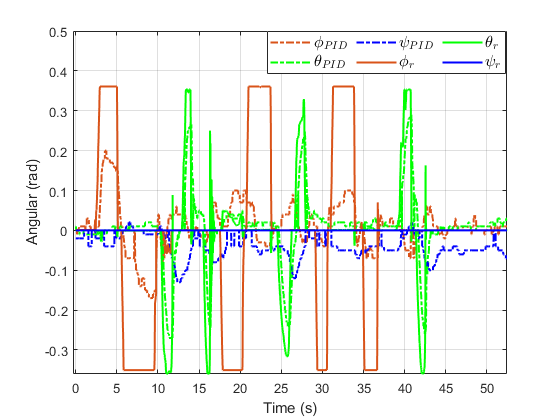}}}
{\resizebox*{4.25 cm}{!}{\includegraphics{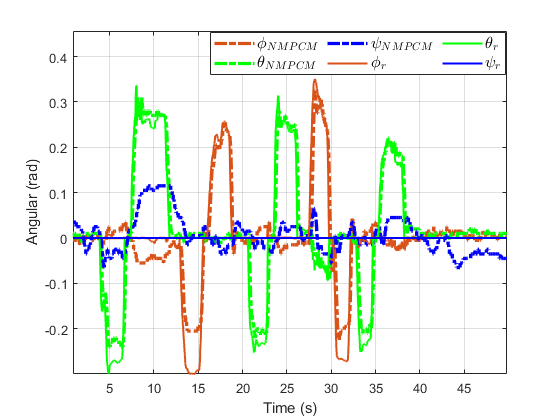}}}
\vspace{-20pt}
\caption{The angular of the Quadrotor UAVs: Cascaded PID (left); NMPCM (right).} 
\label{angularex}
\end{figure}
\begin{figure} [H]
\vspace{-20pt}
\centering
{\resizebox*{4.25 cm}{!}{\includegraphics{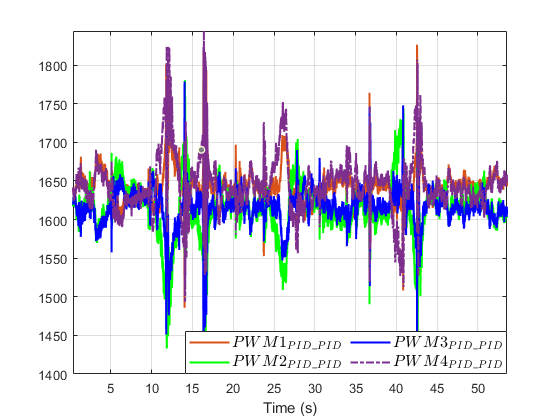}}}
{\resizebox*{4.25 cm}{!}{\includegraphics{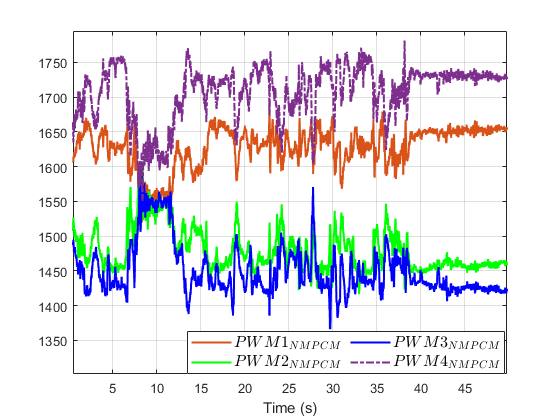}}}
\vspace{-20pt}
\caption{The PWM control signals: Cascaded PID (left); NMPCM (right).} 
\label{pwmex}
\end{figure}
\begin{figure} [H]
\vspace{-20pt}
\centering
{\resizebox*{4.25 cm}{!}{\includegraphics{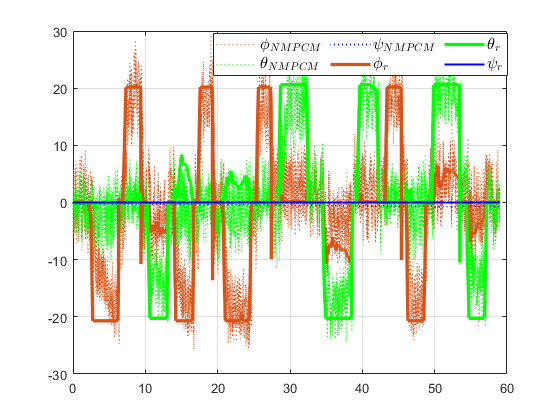}}}
{\resizebox*{4.25 cm}{!}{\includegraphics{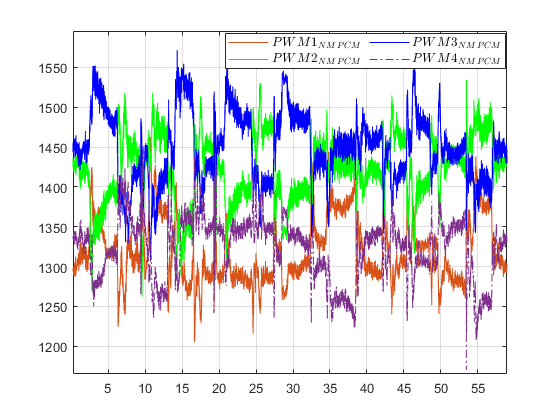}}}
\vspace{-20pt}
\caption{The angular and PWM signals of the H-frame: Angular (left); PWM (right).} 
\label{hframe}
\end{figure}
\begin{figure} [H]
\centering
{\resizebox*{4.25 cm}{!}{\includegraphics{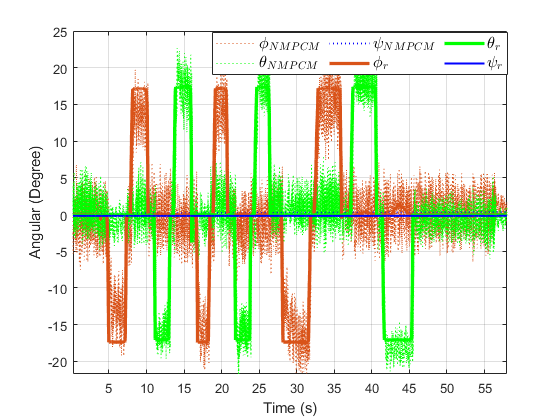}}}
{\resizebox*{4.25 cm}{!}{\includegraphics{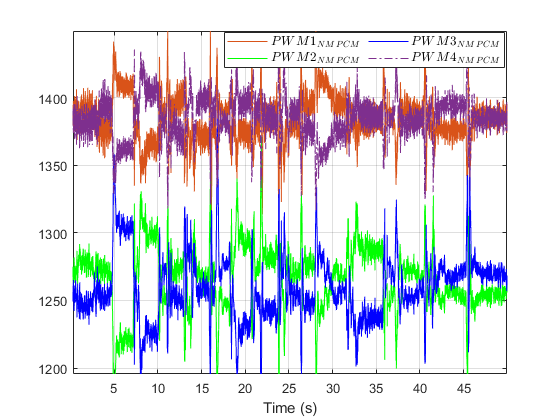}}}
\vspace{-20pt}
\caption{The angular and PWM signals of the DJI-frame: Angular (left); PWM (right).} 
\label{djiframe}
\end{figure}
\begin{figure} [H]
\vspace{-20pt}
\centering
{\resizebox*{4.25 cm}{!}{\includegraphics{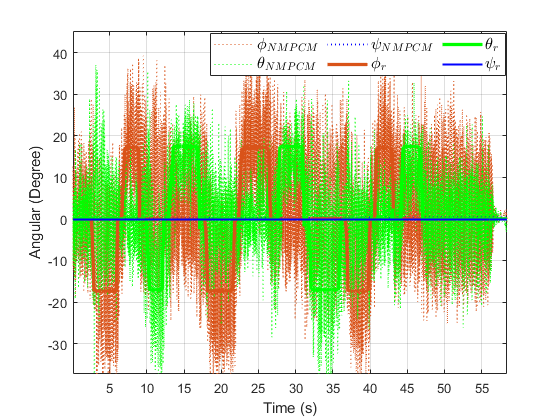}}}
{\resizebox*{4.25 cm}{!}{\includegraphics{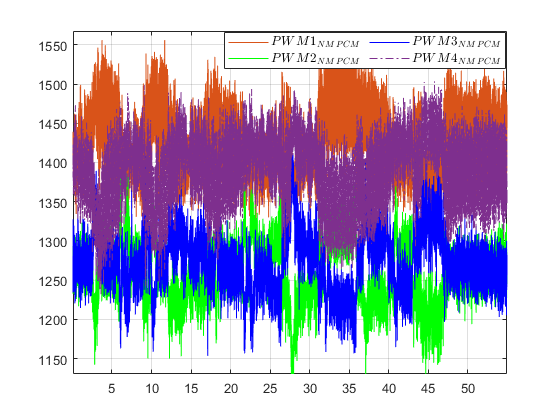}}}
\vspace{-20pt}
\caption{The angular and PWM signals of the X-frame: Angular (left); PWM (right).} 
\vspace{-10pt}
\label{2312motor}
\end{figure}
\begin{table} 
\centering
\caption{The experimental control performance benchmark}
{\begin{tabular}{|l|l|l|l|l|l|l|} \hline
\bf{States}&\bf{Control strategy}&\bf{ISE}& \bf{ITSE}& \bf{IAE}  & \bf{ITAE} \\ \hline
{$\phi$} (rad)   & Cascaded PID & 1.673 & 35.784  & 6.051 & 128.815 \\ 
                 & NMPCM & 0.061 & 1.317  & 1.091 & 24.291  \\  \hline
{$\theta$} (rad) & Cascaded PID & 0.217 & 5.591 & 1.953 & 51.841 \\ 
                 & NMPCM  & 0.052 & 1.081 & 0.999 & 22.785  \\ \hline
{$\psi$} (rad)   & Cascaded PID& 0.129 & 3.765 & 2.129 & 63.611 \\ 
                 & NMPCM  & 0.081 & 1.619 & 1.503 & 36.011 \\ \hline
\end{tabular}}
\label{performancebenmarkafteso}
\end{table}	
\subsubsection{Evaluation of tracking desired}
For the trajectory tracking performance, we send the desired angular values via the radio receiver. The tracking performance of the proposed method compared to the Cascaded PID is shown in Figure~\ref{angularex} - Figure~\ref{2312motor}. Moreover, the control performance benchmark in the experiment is presented in Table~\ref{performancebenmarkafteso}. It can be observed that our control method outperforms the Cascaded PID and provides a smooth PWM signal while tracking the desired trajectory.
\section{Conclusion}
In this paper, we present an embedded NMPC framework designed for resource-constrained microcontrollers, specifically applied to quadrotor UAVs. The Cascaded PID controller provides an initial solution for the optimization problem, while the nonlinear dynamics of the system are addressed using the ACADO code generation tool and the qpOASES solver, which employs the Gauss-Newton method. The embedded code, running on a Teensy 4.1 microcontroller, achieves high-frequency performance, demonstrating its practicality for real-time systems. We evaluate the performance of the proposed method through both simulations and real-time experiments. In the simulations, we include a benchmark comparison with other state-of-the-art methods, such as MPCC (Model Predictive Contouring Control), and advanced solvers like CaSaDi. For real-time experiments, the proposed NMPCM framework not only achieves high-frequency computation on a single microcontroller but also successfully stabilizes the quadrotor UAVs while satisfying input constraints. This highlights the effectiveness and efficiency of the proposed approach in real-world applications. In the future, we plan to incorporate an additional layer using potential field or optimal-time optimization for navigation, which will consider obstacle avoidance and external disturbances to enable the autonomous operation of the system. For more details on embedding the code into the microcontroller and accessing the code, please visit our website: \url{https://github.com/aralab-unr/NMPCM}

\bibliographystyle{IEEEtran}
\balance
\bibliography{RefFile}
\end{document}